\documentclass[lettersize,journal]{IEEEtran}

\usepackage{graphics} 
\usepackage{stfloats}
\usepackage{graphicx}
\usepackage{epsfig}   
\usepackage{mathptmx} 
\usepackage{times}    
\usepackage{amsmath}  
\usepackage{amssymb}  
\usepackage{enumerate}
\usepackage{algorithm}
\usepackage{algorithmicx}
\usepackage{algpseudocode}
\usepackage{adjustbox}
\usepackage{array}
\usepackage{hyperref}
\usepackage{booktabs}
\usepackage{placeins}

\hypersetup{
    colorlinks = true,
    linkcolor  = black,     
    urlcolor   = NavyBlue,
    citecolor  = NavyBlue 
}
\usepackage[table,svgnames]{xcolor}
\usepackage{footmisc}
\usepackage{tabularx}
\usepackage{subcaption}
\usepackage{multirow}

\title{Vision-based Manipulation of Transparent Plastic Bags in Industrial Setups}

\author{F. Adetunji$^{\dagger\ddagger*}$, A. Karukayil$^{\dagger\ddagger*}$, P. Samant$^{\dagger\ddagger*}$, S. Shabana$^{\dagger\ddagger*}$, F. Varghese$^{\dagger\ddagger*}$, U. Upadhyay$^{\dagger\ddagger*}$, R. A. Yadav$^{\dagger\ddagger*}$, \\
A. Partridge$^{\ddagger}$, E. Pendleton$^{\ddagger}$, R. Plant$^{\ddagger}$, Y. Petillot$^{\dagger\ddagger}$, M. Koskinopoulou$^{\dagger\ddagger}$
\thanks{$^{\dagger}$The School of Physical and Engineering Sciences, Heriot-Watt University, Edinburgh, UK. \\
$^{\ddagger}$ The National Robotarium, Edinburgh UK. \\
Corresponding author email: m.koskinopoulou@hw.ac.uk.} 
\thanks{$^{*}$ Authors contributed equally to this work.}
}

\begin{document}

\maketitle

\begin{abstract}
This paper addresses the challenges of vision-based manipulation for autonomous cutting and unpacking of transparent plastic bags in industrial setups, aligning with the Industry 4.0 paradigm. Industry 4.0, driven by data, connectivity, analytics, and robotics, promises enhanced accessibility and sustainability throughout the value chain. The integration of autonomous systems, including collaborative robots (cobots), into industrial processes is pivotal for efficiency and safety. 
%
The proposed solution employs advanced Machine Learning algorithms, particularly Convolutional Neural Networks (CNNs), to identify transparent plastic bags under varying lighting and background conditions. Tracking algorithms and depth sensing technologies are utilized for 3D spatial awareness during pick and placement. The system addresses challenges in grasping and manipulation, considering optimal points, compliance control with vacuum gripping technology, and real-time automation for safe interaction in dynamic environments. The system's successful testing and validation in the lab with the FRANKA robot arm, showcases its potential for widespread industrial applications, while demonstrating effectiveness in automating the unpacking and cutting of transparent plastic bags for an 8-stack bulk-loader based on specific requirements and rigorous testing.
\end{abstract}

\begin{IEEEkeywords}
autonomous system, industrial applications, vision-guided manipulation, transparent bag detection and manipulation
\end{IEEEkeywords}

\section{INTRODUCTION}
Industry 4.0—also called the Fourth Industrial Revolution or 4IR—is the next phase in the digitization of the manufacturing sector, driven by disruptive trends including the rise of data and connectivity, analytics, human-machine interaction, and improvements in robotics \cite{edel2022,zhong2017intelligent}.
This could make products and services more easily accessible and transmissible for businesses, consumers, and stakeholders all along the value chain \cite{Goel2020}. Preliminary data indicate that successfully scaling 4IR technology makes supply chains more efficient and sustainable \cite{JAVAID202158}, creates a safer and more productive environment for the employees, reduces occupational accidents and factory waste, and has countless other benefits.

Autonomous manipulation of plastic packages in industrial setups typically involves the use of robotic systems and automation technologies \cite{ALI2018205}. These systems are designed to handle, move, and manipulate plastic packages in a variety of industrial processes, such as packaging, recycling and sorting, food processing, and quality control \cite{Gabellieri2023, Chen2023}. 

Collaborative robots, or cobots, are widely used in various industrial applications, working alongside humans without needing extensive safety barriers, cages, or other restrictive measures \cite{chen2018}. These robots use different sensors to identify their environment, recognise objects and are programmed for better accessibility, flexibility and repeatability. Example cases can be found in the textile industry as described in \cite{Kragic2021}, where the authors proposed a dual arm collaborative system for textile material identification. By imitating human behavior, in this work the robots use actions such as pulling and twisting to identify and learn more about textile properties. In recent years, the recycling and waste management industry has begun to use vision-based robotic systems for the classification and accurate sorting of waste materials \cite{zenRob}. Indicative examples can be found in different recycling industries for the management of construction waste \cite{CHEN2022,zenArticle}, recyclable materials \cite{mkosk2021,epapad20} or electronic parts \cite{PARAJULY2016652,Naito2021}.

The vision-based manipulation and autonomous cutting of transparent plastic bags presents a set of intricate challenges and a compelling need for innovative AI solutions \cite{BillardKragic2019, TIAN2023}. The inherent transparency of the bags poses difficulties in accurate detection due to the reflection and refraction of light, demanding sophisticated computer vision algorithms for reliable identification \cite{SajjanICRA2020}. 
The deformable nature of plastic bags adds complexity to the grasping and manipulation process, necessitating advanced robotic control strategies to handle their variability \cite{MAKRIS2023513}. 

Additionally, autonomous cutting requires well-considered mechanical design and precise vision-guided tools to discern optimal cutting points while avoiding unintended damage. Ensuring the safety and efficiency of these systems in real-time, dynamic environments further amplifies the challenge. The pressing need for such technologies arises from the increasing demand for automated waste management, recycling, and packaging processes, where vision-based systems can enhance efficiency, reduce human intervention, and contribute to sustainable practices by facilitating the effective processing of transparent plastic bags \cite{Ahmad2023}.

%
In this work, through the use of advanced Machine Learning algorithms, based on Convolutional Neural Networks (CNNs), the system can identify transparent plastic bags within its visual field, taking into account variations in lighting and background. Once the bags are detected, the system utilizes tracking algorithms to follow the pick and placement of the bags, and, integrate depth sensing technologies for 3D spatial awareness. The next steps involve developing algorithms for robotic grasping and manipulation, accounting for the challenges posed by the deformable and transparent nature of plastic bags. This includes considerations for optimal grasping points, compliance control using vacuum gripping technology, and real-time automation and processing to ensure effective and safe interaction with the bags in dynamic environments. 

The rest of the paper is organized as follows. Section II describes the mechanical design of the proposed system. Section III presents the object detection and manipulation approach based on deep-learning and Section IV presents the autonomous cutting mechanism and the automation process. 
The testing of the pilot proof-of-concept prototype is presented in Section V.
Finally, the last section discusses the obtained results and highlights directions for future work.

\section{MECHANICAL DESIGN}

The mechanical design encompasses three key components:

\begin{enumerate}[i]
    \item Feeding: This involves the precise picking and placing of eight packaged plastic-container stacks from an adjacent tote into eight individual enclosures aligned with the bulk loader's feeding system.
    \item Cutting: This entails the opening, removal, and disposal of the packaging surrounding the plastic-container stacks. This operation is performed while the plastic-container stacks are securely held within the eight individual enclosures.
    \item Delivery: This stage involves opening the enclosures containing the plastic-containers and strategically placing the unpackaged stacks into the bulk loader, facilitating the seamless integration of the plastic-containers into the larger industrial process.
\end{enumerate}
This section details each subsystem within the design of the prototype.\\

\begin{figure}[h]
    \centering
    \includegraphics[width=0.95\columnwidth]{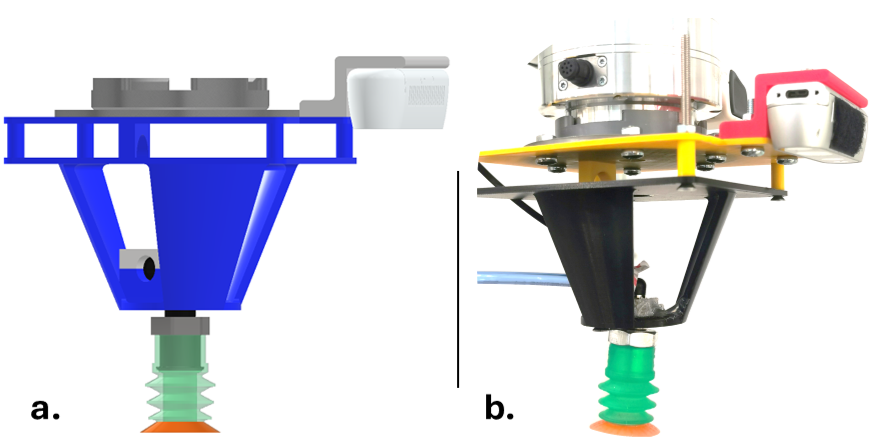}
\caption{a. CAD model of the vision-based gripping system; b. real system embedded in the Franka robot.}
\label{fig:camera_sys}
\end{figure}

\begin{figure*}[h]
    \centering
    \includegraphics[width=\textwidth]{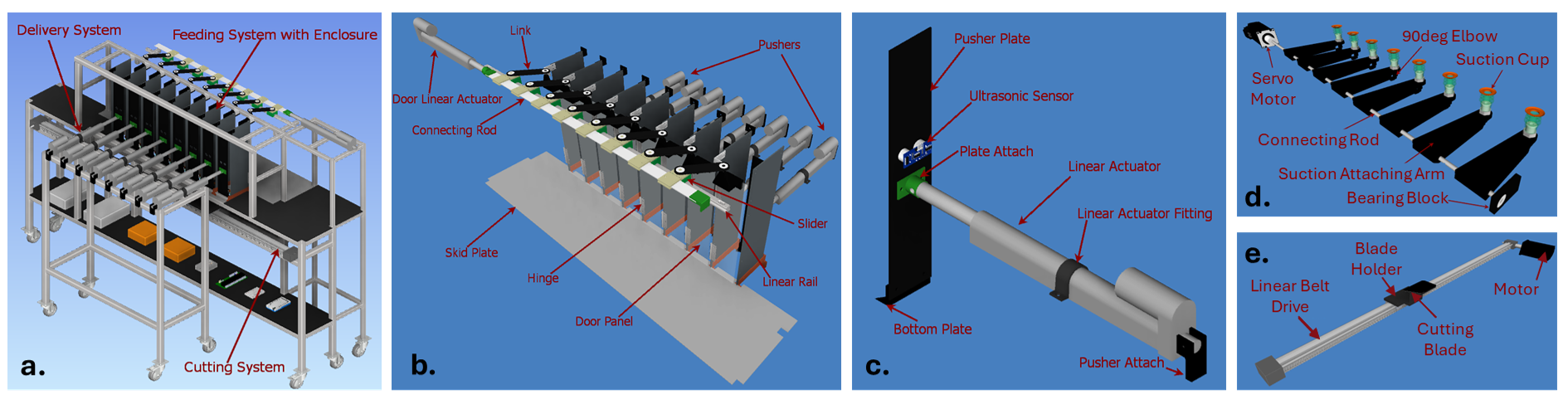}
\caption{CAD model designs of the system's components: a. eight-stack design assembly; b. delivery system; c. pushing mechanism; d. bottom suction mechanism; e. cutting mechanism.}
\label{fig:mechan_components}
\end{figure*}

\noindent
\textbf{Feeding.}
In the feeding system, a collaborative robot arm (the 7-axis Franka Emika Panda, equipped with an Intel RealSense D350 camera and a custom suction cup gripper) has been employed for manipulation and vision tasks (Figure \ref{fig:camera_sys}b). The gripper comprises a Schmalz PSPF 33 SI-30/55 G1/8-AG suction cup, an SBP 15 G02 SDA vacuum generator, and a VS VP8 SA M8-4 pressure sensor enclosed in a custom 3D printed housing (Figure \ref{fig:camera_sys}a: CAD model of the custom gripper and b:real printed griping system).
The feeding process initiates with the camera capturing a top-down image of the tote, identifying the tops of the plastic stacks, and assigning a value to establish the picking order. The robot arm, guided by the established order, uses suction to pick and place individual stacks into 1 of 8 custom enclosures made of aluminum extrusion and acrylic (Figure \ref{fig:mechan_components})a. Stack placement is verified using HC-SR04 ultrasonic sensors on the enclosure's back wall.
To facilitate picking, the tote containing the stacks is inclined at an angle (12$^{\circ}$) to prevent the stacks from toppling. The tote is labeled with a QR code for arm position estimation between picks. Solenoids control suction activation, and the pressure sensor provides feedback to confirm successful suction. This comprehensive setup ensures effective and reliable feeding for the subsequent stages of the automated unpacking and cutting process. \\

\noindent
\textbf{Cutting.}
The cutting system consists of two interconnected components: a gripping mechanism and a cutting mechanism. The gripping mechanism employs eight vacuum-driven suction cup grippers to securely hold the bottom of each stacks packaging. A vacuum line, directed through two compressors and solenoid valves, divides the vacuum between the cobot arm's end effector gripper and the eight suction cup grippers. Each of the eight individual air conduits is equipped with a dedicated vacuum generator, generating ample vacuum for secure suction, along with pressure sensors to ascertain suction power. Suction cups, mounted on 3D printed arms and connected to motors, swing into contact with the stack bases (Figure \ref{fig:mechan_components}d).
The cutting mechanism features a scalpel blade housed within a custom 3D printed mount, affixed to a linear belt drive (Figure \ref{fig:mechan_components}e). The cutting process commences by opening the solenoid valve, supplying pressure to the vacuum generators. The swinging rod engages each suction cup gripper, making contact with the bottom of all eight packages. As suction secures the packages, the sensors on each gripper gauge the required suction power. Once optimal conditions are reached, the rod rotates, creating tension in the packaging. Simultaneously, the cutting mechanism traverses the linear rail, slicing through the packaging. Upon completion, the valves close, releasing the cut plastic into a container beneath the aluminum frame.
After opening all eight stacks, the cobot arm reverses the pick-and-place task, using the suction cup gripper to grasp the top of each remaining packaging item. These are then placed into a designated disposal bin. This comprehensive cutting system ensures precise and efficient packaging removal, complementing the overall automated process for unpacking and cutting transparent plastic bags in industrial setups. \\

\noindent
\textbf{Delivery.}
The delivery system incorporates nine WL-22040921 linear actuators, with eight arranged in parallel through a two-channel relay module to create the pushing mechanism against the back walls of the enclosures. The remaining actuator, also connected to a two-channel relay module, is dedicated to the custom door mechanism. In the pushing mechanism (Figure \ref{fig:mechan_components}c), each linear actuator, positioned against the back walls, executes forward movements, serving as pushers, and integrates ultrasonic sensors for precision. A custom plate at the base catches the stacks, facilitating their smooth displacement into the bulk loader.
For the custom door mechanism (Figure \ref{fig:mechan_components}b), a linear guide rail, linear actuator, door hinges, and acrylic doors are utilized. Eight individual sliders on the rail correspond to the linear actuator, doors, and hinges, enabling synchronized opening and closing. A limit switch at the rail's end prevents excess movement, safeguarding the door mechanism. The delivery process initiates after the cobot arm removes all packaging, with the linear actuator autonomously opening the doors, and the eight linear actuators pushing the unpackaged stacks onto an acrylic skid plate. Upon successful placement, the linear actuators revert to their initial position, retracting the enclosure back walls, and closing the doors.
Upon completion, the system undergoes a reset for the subsequent delivery cycle. This intricately designed delivery system ensures efficient and controlled movement of unpackaged stacks, contributing to the seamless integration of the transparent plastic bag unpacking and cutting process in industrial setups.

\section{DEEP-LEARNING-BASED OBJECT DETECTION AND MANIPULATION}

The vision system's comprehensive workflow for real-time detection and tracking of transparent bags is presented in the following. The camera framework operates on ROS Noetic, leveraging the ROS Wrapper for Intel RealSense Devices provided by Intel. By initiating the RealSense camera package, the camera commences the publication of depth and vision (RGB) data, readily accessible for subscription and utilization as needed. These captured data are transformed into monochrome and disseminated to subsequent detection stages. QR codes are employed for zone categorization, aiding in depth data estimation. The detection process is executed using YOLOv5, integrated into ROS through the ROS wrapper. YOLOv5 is renowned for its efficiency and performs real-time detection of plastic-container stacks, providing precise picking locations to the robotic system. 

\subsubsection{QR Code Based Depth Estimation and Tracking}
\begin{figure}[h]
    \centering
    \includegraphics[width=0.32\columnwidth]{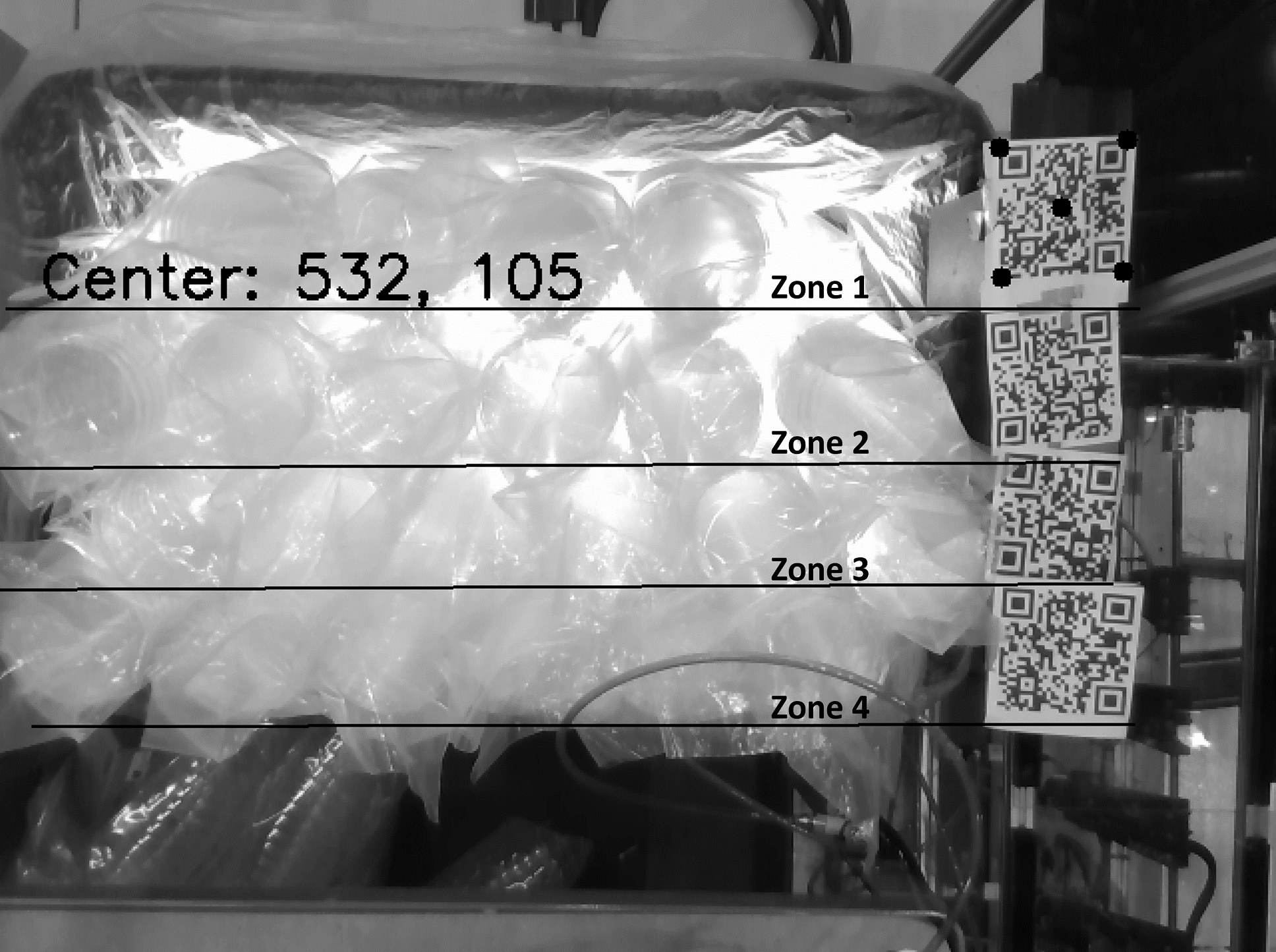}
    \includegraphics[width=0.32\columnwidth]{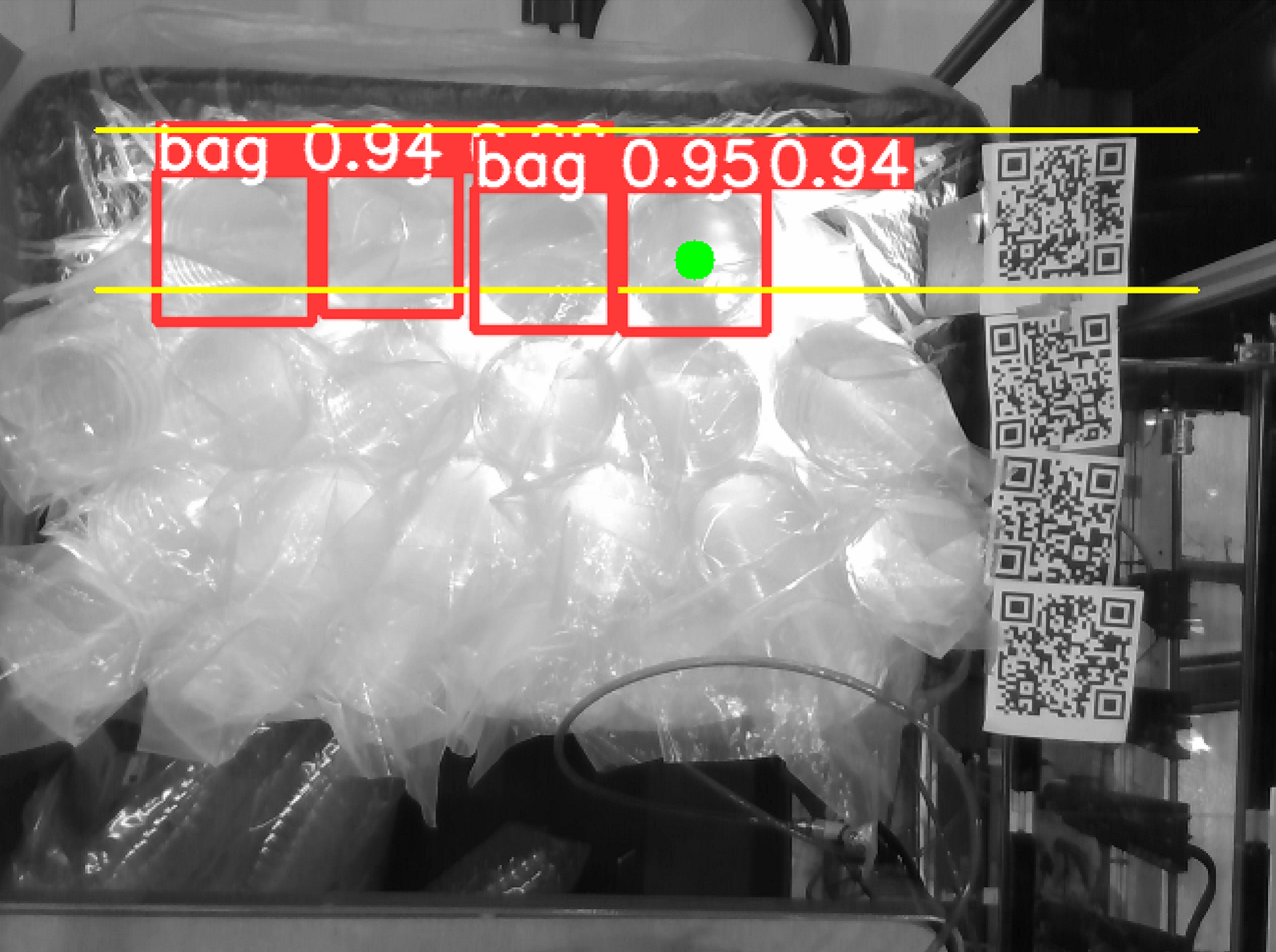}
    \includegraphics[width=0.32\columnwidth]{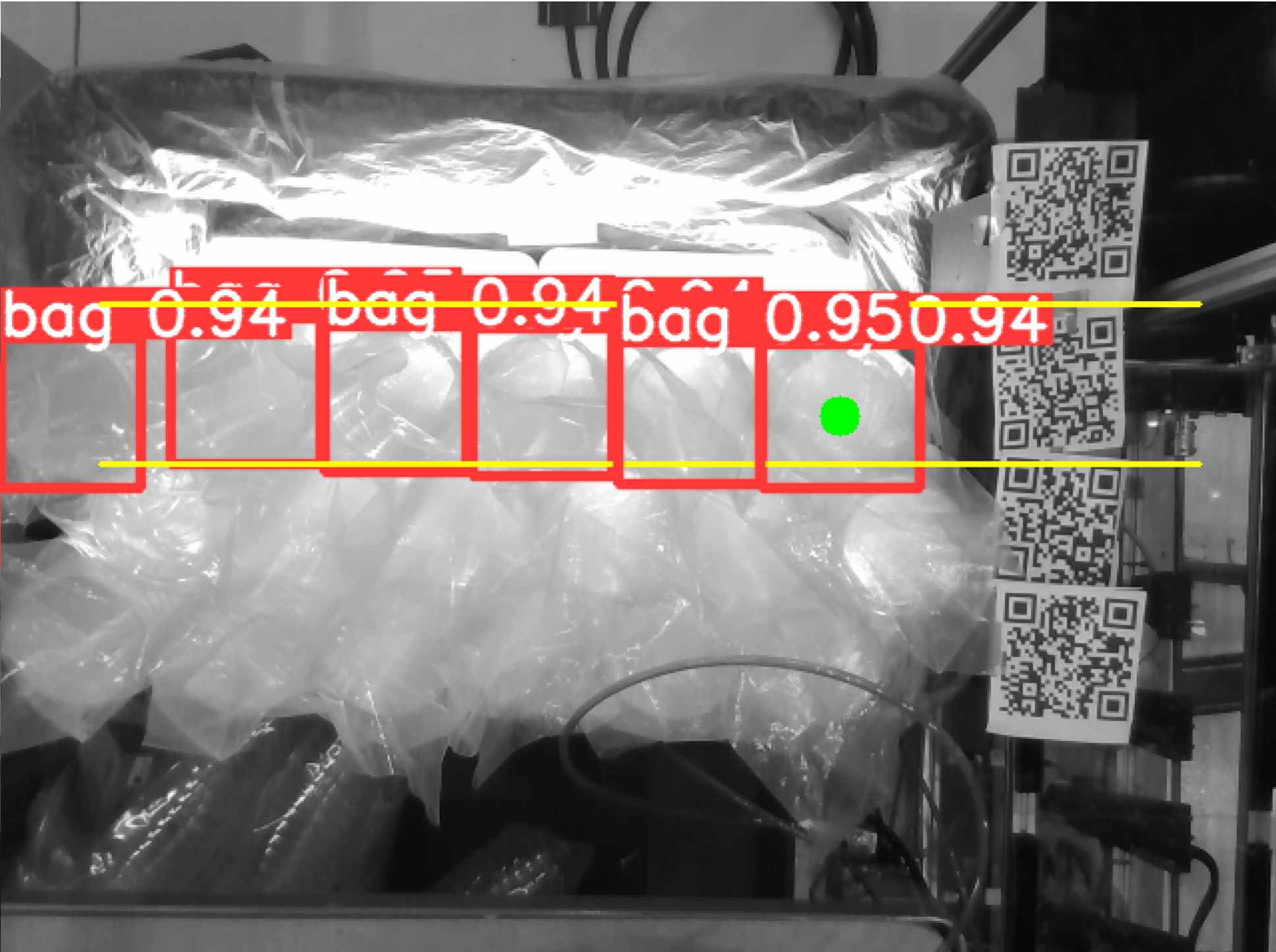}
\caption{Plastic-container stack detection process. (a) Zone identification and QR-code detection of zone 1, along with its center co-ordinates; (b) First zone stack detection and tracking of the next picking target (marked with the green dot); (c) plastic-container stack detection of zone 2.}
\label{fig:QRdetection}
\end{figure}

Depth estimation is required for the conversion of the camera coordinates to Cartesian coordinates. However, this is non-trivial as depth estimation using the current camera is not always robust due to inconsistencies caused by the transparency of the bags and plastic-containers. To overcome this, QR codes have been placed at the side of the tote as shown in Figure \ref{fig:QRdetection}a. The QR codes provide clear points of reference and an accurate depth reading at the beginning of the task. Four QR codes have been used in total, each of which corresponds to one of the four rows of stacks in the tote. The distance between the stack and the camera is then calculated based on the distance between the camera and the respective QR code. This also allows the bags to be grouped into different zones as shown in Figure \ref{fig:QRdetection}b. Picking of the stacks presupposes optimal tracking, such that the robot can return to the next stack in the sequence after loading the feeding system. In order to achieve this, the detection of the bags is performed by zone rather than by tote. In Figure \ref{fig:QRdetection}b, the detection of the stacks of zone 1 is shown, by identifying 4 bags in red and their corresponding confidence level. The green dot within the right-hand red box indicates the target stack that the robot is going to pick next. Whereas, Figure \ref{fig:QRdetection}c illustrates the process sequence after the first zone stacks are successfully picked and placed by the robot and the detection of the 6 stacks of zone 2.
By finding the coordinates of the leftmost stack, the robot can detect and pick the stacks one by one from left to right. Once the picking of the stacks of zone 1 has been completed, it can then proceed to zones 2, 3 and finally 4.

\subsubsection{Data Acquisition} 
\label{Data acquisition and annotation}
\begin{figure}[h!]
    \centering
    \includegraphics[width=\columnwidth]{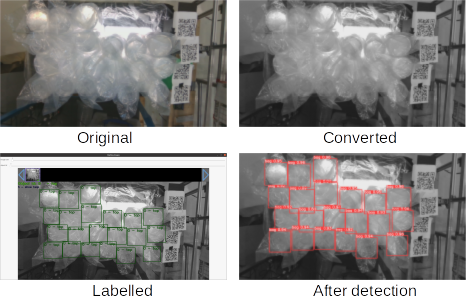}
    \caption{YOLO\_V5 training process: \emph{Original} is the captured image; \emph{Converted} is the grayscale image; \emph{Labelled} is the ground truth; and the last one is the result \emph{after detection}.}
    \label{fig:YOLOV5 training process}
\end{figure}

A custom data-set was created using the same Intel D435 camera used for object detection by capturing the plastic-container stacks and labelling them to be used as training data for the algorithm. A sample image from the raw images captured is shown in Figure \ref{fig:YOLOV5 training process} (\emph{Original}). The images were grouped into training, validation and test sets of 150, 40 and 24 images respectively. The data were collected under different environmental lighting and across various time periods within this project. The 24 images in the test dataset were captured during experimentation, while emptying the tote by picking the stacks one by one. The camera parameters used for data collection in \href{https://obsproject.com/}{OBS Studio} are listed in Table \ref{tab:camera_settings}.

\begin{table}[h!]
    \centering
    \caption{Camera parameters}
    \begin{tabular}{l l}
        \textbf{Parameter} & \textbf{Value} \\ 
         Resolution & 1920 x 1080 \\ 
         Brightness & 0 \\ 
         Contrast & 50\\  
         Saturation & 64 \\ 
         White Balance Temperature & 4600 \\ 
    \end{tabular}
    \label{tab:camera_settings}
\end{table}

The raw images are then converted to monochrome to reduce noise. The sharpness and contrast were also adjusted to obtain better results. Figure \ref{fig:YOLOV5 training process} (\emph{Converted}) shows the results after this step of image processing.

The dataset was prepared and labelled using the annotation tool \emph{YOLO Mark}\footnote{\href{https://github.com/AlexeyAB/Yolo_mark}{https://github.com/AlexeyAB/Yolo\_mark}} as shown in Figure \ref{fig:YOLOV5 training process} (\emph{Labelled}). After testing multiple offline labelling tools, YOLO\_mark provided reliable results. 

To train the model with the \emph{YOLO\_V5}\footnote{\href{https://github.com/ultralytics/yolov5}{https://github.com/ultralytics/yolov5}} algorithm for object detection, we used the data acquired during the data acquisition process (Section \ref{Data acquisition and annotation}). The training can be done using either the local machine if the local machine has a sufficient NVIDIA GPU or by using the Google Collab Cloud GPU. For the scope of this work Google Collab was used. 

\subsubsection{Robot control for pick-n-place of detected plastic bags} 
\begin{figure}[h]
    \centering
    \includegraphics[width=0.8\columnwidth]{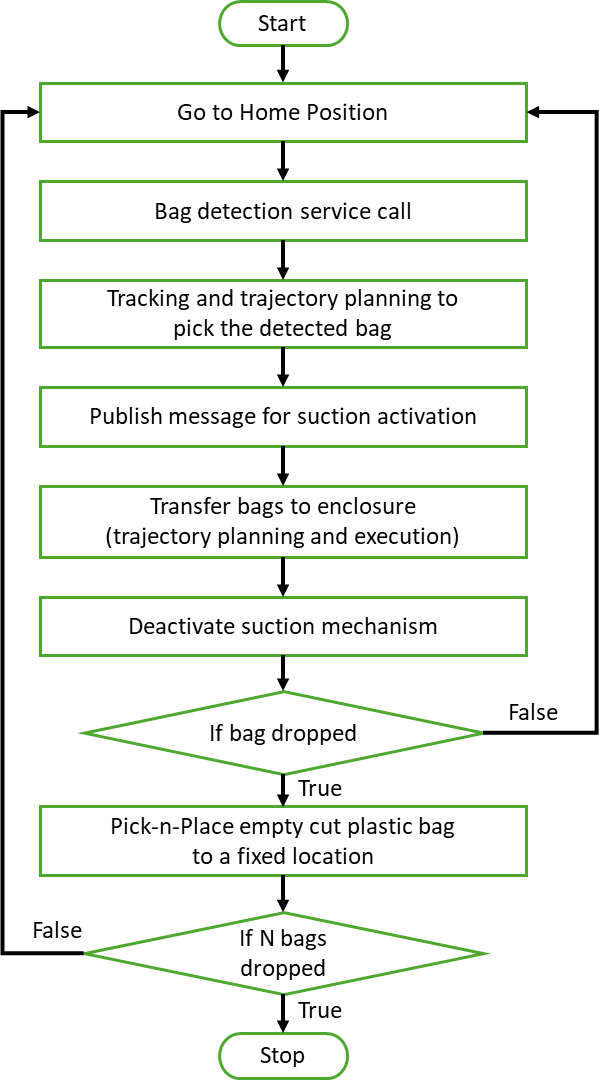}
    \caption{Robotic manipulation workflow.}
    \label{fig:RobotControlWorkflow}
\end{figure}

The robotic manipulation process begins with a fixed \emph{Home Pose}, where the robot is positioned at a predefined location that provides a complete view of all the packaged stacks within the tote. This home position serves as the starting point for subsequent operations.
Next, the robot initiates the detection of packaged stacks which marks commencement of the third workflow step, by invoking the packaged stack detection code through the ROS architecture using a ROS service. The outcome of this service query is the identification of the next single packaged stack to be picked by the robot within the camera's visual frame.  
Figure \ref{fig:RobotControlWorkflow} presents the overall automation logic of the robot control for this pick-n-place task.

Upon successfully detecting the packaged stack, the three-dimensional coordinates of the detected stack in the camera's optical frame are transformed into the robot's reference frame using MoveIt hand-eye calibration package \cite{moveit_calib}. This calibration process generates a calibration file specific to the robotic setup which calculates the cobot configuration from the camera's three-dimensional coordinates.  

The cobot then initiates the fourth workflow step which involves trajectory planning and execution.
The robot uses the integrated motion planner, MoveIt Pilz Industrial Motion Planner (LIN), to plan and execute a linear path to reach the identified packaged stack. 
LIN utilises the Panda Franka Emika's cartesian constraints to create a trapezoidal velocity profile in Cartesian space for the cobot's movement. This profile ensures the cobot accelerates, maintains a constant speed, and then decelerates during its movements. This approach proves highly effective, especially when handling packaged stacks as they can deform after collisions with objects in the workspace if the speed profiles are not controlled. 

To further refine control, scaling factors for Cartesian velocity and acceleration are integrated in the system, which imposes a maximum speed limit on the trajectories generated by the planner. The speed and planning parameters used for the robots pick-n-place testing are tabulated in Table \ref{tab:cobots-parameters}.

\begin{table}[h!]
    \centering
    \caption{Cobot Speed and Planning Parameters}
    \begin{tabular}{l l}
        \toprule
        \textbf{Parameter} & \textbf{Value} \\
        \midrule
         Max. Cartesian Speed of Franka Robot &  2 m/s \\ 
         Max. Set Velocity Scaling Factor & 0.28 \\ 
         Max. Set Acceleration Scaling Factor & 0.03 \\ 
         Max. Cartesian Speed & 0.56 m/s \\ 
         Max. Planning Time (Motion Planner) & 5 s \\ 
         Max. No. Planning Attempts (Motion Planner) & 10 \\ 
        \bottomrule
        \label{tab:cobots-parameters}
    \end{tabular}
\end{table}

Following the workflow pipeline of Fig.\ref{fig:RobotControlWorkflow}, after target position reached, the robot publishes a message to the ROS framework to activate the suction mechanism and grasp the identified packaged stack. The robot then transfers the stack to an empty enclosure in the cutting module completing the sixth workflow step.  This is accomplished using the motion planner to plan a path through a set of predefined waypoints which ensures collision avoidance whilst navigating through the workspace.

After reaching the specified enclosure, the cobot begins the next process workflow step. Here, the robot utilises the ROS framework to publish another message to the suction node. This triggers the node to cut off the suction supply, thereby releasing the packaged stack to place it safely within the enclosure. 
This entire process is repeated for a further seven packaged stacks to fill the eight cutting module enclosures, following the eight-step workflow which loops previous steps until the decision block is true. The robot carries out the sequence, ensuring that each stack is picked up, transferred, and released with precision.

In turn, the cutting mechanism is activated and the robot is used to retrieve the cut plastic bags. The robot utilises the Pilz motion planner to plan and execute paths through sets of predefined waypoints to move the cobot from above the enclosures to a designated bin for disposing of the cut plastic bags, this entails the ninth step in the workflow. The end effector is positioned above the enclosure and the suction activated such that it grasps the top of the cut plastic packaging. The packaging is removed from the stack by a vertical upward movement of the cobot end effector. The robot then moves its end effector to above the designated bin, and the suction node is invoked to deactivate the suction mechanism and release the cut plastic bag, dropping it into the bin. This process is repeated for all eight packaged stacks in the enclosures.  The tenth step in the workflow involves repeating the above actions until all twenty-four plastic bags are cut and disposed of, thus ending the workflow.

\section{AUTONOMOUS CUTTING \& CONTROL}

The whole system is integrated with the aid of robotic control and electronic automation. The robotic control is developed in the Robot Operating System (ROS), through customisation of libraries such as motion-planner and move-it. 
A custom made automation solution oversees the whole operation through feed-back from various sensors and the electronic actuation of motors. The automation logic is implemented using a Raspberry-Pi module which acts as the master controller and commands the Arduino module, which in turn handles the actuation and feed-back based on the commands sent by the master module.

The automation system uses a combination of Arduino Mega, sensors (including ultrasonic and pressure sensors), linear actuators, solenoids, and a Raspberry Pi 4 for automation. The sensors and actuators are linked to the Arduino Mega. The Arduino communicates with the Raspberry Pi 4 via serial communication. The Raspberry Pi 4 reads sensor data transmitted by the Arduino and sends commands to activate solenoids, motors, and linear actuators. 

The Raspberry Pi is coded using ROS python which is the automation controller in this implementation. It coordinates actions with the cobot by publishing and subscribing to the relevant topics at the appropriate time. 

The operation starts with the Raspberry Pi controller publishing system ready to send commands to the cobot, which then initializes the cobot operation. The cobot then moves to a position over the tote, so that it can detect the stacks, and gives the ready for picking command to the controller. The Raspberry Pi then switches on the solenoid valve connected with the cobot suction gripper, which aids the picking of the bags from the tote. The cobot then approaches the bag for picking, meanwhile the controller monitors the pressure sensor value from the cobot gripper to check whether the pick has been successful. If the picking fails the cobot moves to the reset position and restarts the picking process. If the pick is successful, the cobot moves to the dropping position over the enclosure. The cobot then gives the drop command to the pi module, indicating its position over the dropping zone. The controller then activates the bottom suction solenoid valves and rotates the motor to position the suction cups below the stacks. The bottom suction will remain in this state for the rest of the cycle. 

Feedback from the ultrasonic sensor is checked to ensure the stack's position within the enclosure. If the stack is identified, feedback is given to the cobot, if not, the cobot will move back to drop position again and reattempt the placement of the stack.  The pressure is checked during this time using the pressure sensors, and if it drops below a cut off value and the ultrasonic sensor still detects a stack, feedback is given which triggers the cobot to start the next feeding operation to the adjacent enclosure position. If the bottom suction pressure drop is not detected within the scheduled time, a failed state is fed back to the cobot, which moves back to the drop position to repeat the dropping again. This cycle is continued until all eight stack positions in the enclosure are filled with the bags. Once all 8 positions are filled, a finished cycle message is received from the cobot controller which then triggers the bottom suction motor to rotate to create tension on the bag. Cutting is done using the motor control interface separately. The automatic control of the system is paused until these operations are completed. 

Once cutting is complete, the Raspberry controller publishes to start the removal of the bags, along with the turning off of the bottom suction and rotation of the suction cups back to their home position. The cobot then moves above the enclosures to remove the packaging, and publish a ready for removal command to the controller. The controller then activates the cobot suction and starts monitoring the pressure sensor value. The cobot then moves towards the stack to pick the bag from the enclosure. The cobot picks the bag and moves to a safe drop position, where it publishes a `removing bag' status check message, which enables the Raspberry Pi to turn off the cobot suction to release the bag.

\begin{figure}[h]
    \centering
    \includegraphics[width=\columnwidth]{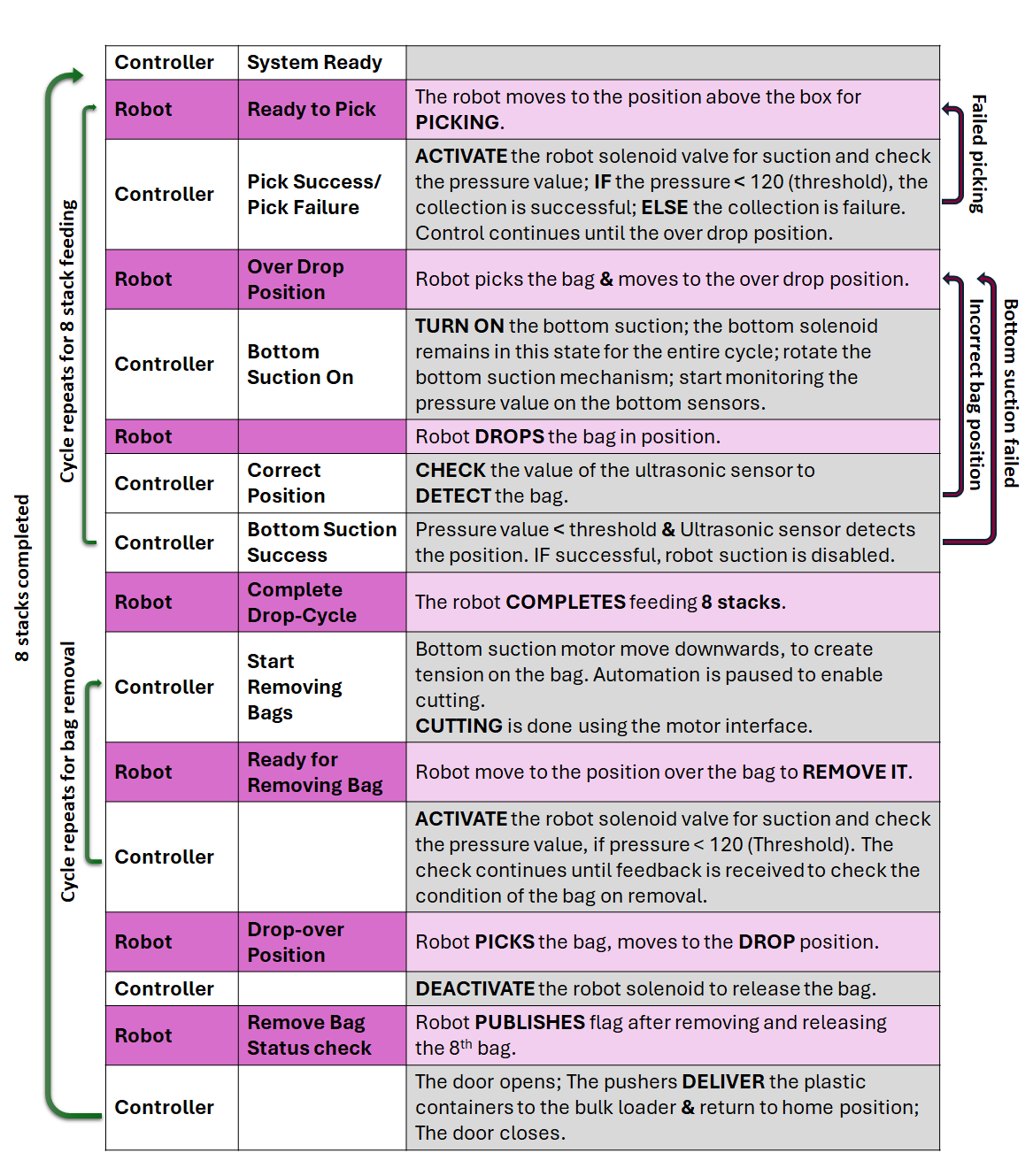}
    \caption{Schematic representation of automation logic.}
    \label{fig:automation}
\end{figure}

This cycle has been repeated for all 8 stacks, and then the pushing mechanism starts, with the doors swinging open. Then the pushers are activated which push the unpacked plastic-containers into the bulk loader over the skid plate. Once this delivery is completed the pushers move back to their home position and the doors close. This completes one cycle of operations for all of the 8 stacks.
The system is then reset by publishing system ready by the rasberry-pi controller to continue the operation for the next feeding cycle.
A simplified representation of the automation logic is shown in Figure \ref{fig:automation}.

\section{EXPERIMENTAL RESULTS}

\begin{figure*}[h!]
    \centering
    \includegraphics[width=\textwidth]{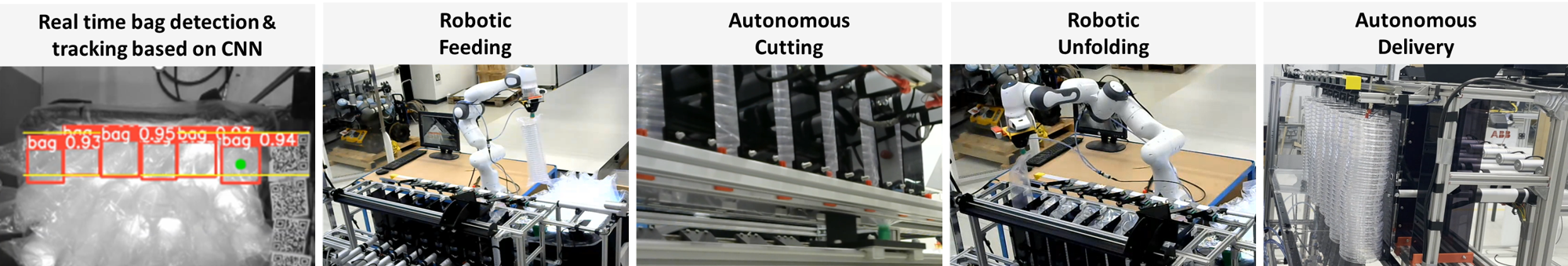}
\caption{Experimental procedure with five identified phases: a. real time bag-detection and tracking; b. robot feeding; c. autonomous cutting; d. robot unfolding and e. autonomous delivery.}
\label{fig:main_procedures}
\end{figure*}

To evaluate the performance of our proposed vision-based manipulation system for the autonomous cutting and unpacking of transparent plastic bags, we conducted a series of experiments consisting of 10 iterations of the complete cycle. Each cycle involved five distinct phases as shown in Fig.\ref{fig:main_procedures}: real-time bag detection and tracking, robot feeding, autonomous cutting, robotic unfolding, and autonomous delivery to the bulk loader. The following sections detail the outcomes and observations for each phase.\\

\noindent {\bf i. Real-Time Bag Detection and Tracking: Vision Detection Performance.}
The first phase involved the detection and tracking of transparent plastic bags using Convolutional Neural Networks (CNNs). The system demonstrated high accuracy in identifying the bags under varying lighting and background conditions. Across the 10 iterations, the average detection accuracy was 96.8\%, with a standard deviation of 1.2\%. The tracking algorithm maintained a robust performance, ensuring continuous monitoring of the bags' positions with an average tracking error of 1.5 mm.

To assess the performance of the trained network we followed the standard evaluation procedure considering three metrics, namely (i) precision, (ii) recall, and (iii) F1-score, which are calculated as follows:

\begin{equation}
\mathit{precision} =  \frac{tp}{tp+fp}
\label{eq:map}
\end{equation}

\begin{equation}
\mathit{recall} =  \frac{tp}{tp+fn}
\label{eq:mar}
\end{equation}

\begin{equation}
\mathit{F1-score} = 2* \frac{precision\cdot recall}{precision + recall}
\label{eq:f1score}
\end{equation}

In these equations tp, fp and fn denote respectively the true positive, false positive and false negative identifications of the plastic bags.
true positives were considered for the cases when
the predicted and real bounding box pair has an IoU score that exceeds the imposed threshold IoU=0.5.
Table \ref{tab:yolov5results} summarises the results of the bag detection performance on both the validation and test set. As the model has been trained on a targeted dataset acquired from the same environment, the inference results very high scoring on average 100\% accuracy to all the experiments conducted.


\begin{table}[h!]
    \centering
    \caption{YOLO\_V5 validation results on the test dataset}
    \begin{tabular}{l l l l}
    \toprule
         \textbf{Precision} &  \textbf{Recall} &  \textbf{F1-score} &\textbf{mAP@0.5} \\
         \midrule
         \multicolumn{4}{c}{Validation Set}\\
         99.5\% & 98.7\% & 99.1\%& 99.4\% \\
         \midrule
         \multicolumn{4}{c}{Test Set}\\
         100\% & 100\% & 100\% &99.5\% \\
         \bottomrule
    \end{tabular}
    \label{tab:yolov5results}
\end{table}

\noindent {\textbf{ii. Robot Feeding.}}
In the second phase, the FRANKA robot arm picked the bags one by one from the box and placed them into each enclosure of the feeding system until all eight enclosures were filled. Table IV provides numerical counts for the number of stacks detected, picked, and placed, with the maximum result for each being eight. The average success rates for picking and placing were 86.25\% and 82.5\%, respectively, with an overall average of seven out of eight bags successfully handled. The average time taken for the robot to complete this task was 8.3 minutes per iteration, with a standard deviation of 0.5 minutes. Challenges in this phase included the suction system’s grip failures and workspace constraints, leading to collisions and placement errors. Overall, as tabulated in Table \ref{tab:Pick_and_place}, the average numbers of successful picked and placed is 7 out of 10. \\

\noindent {\textbf{iii. Autonomous Cutting.}}
During the third phase, the autonomous cutting mechanism was activated to cut all eight bags. The cutting process was highly efficient, with an average completion time of 15.7 seconds per iteration and no cutting errors observed across all 10 iterations. The precision of the cuts was within an acceptable margin, with an average deviation of 0.2 mm from the intended cut lines. This phase did not experience the interdependent issues observed in detection, picking, and placing, thus maintaining consistent performance.\\

\noindent {\textbf{iv. Robotic Unfolding.}}
The fourth phase involved the robotic unfolding of each of the eight bags. The FRANKA robot arm demonstrated high dexterity and control, successfully unfolding all bags in each iteration. The average time taken for unfolding all eight bags was 38.9 seconds, with a standard deviation of 2.8 seconds. The system's compliance control with vacuum gripping technology ensured minimal damage to the bags and plastic containers during the unfolding process.\\

\noindent {\textbf{v. Autonomous Delivery to Bulk Loader.}}
In the final phase, the unfolded bags were autonomously delivered to the bulk loader by activating the pushers. The system successfully delivered all eight bags in each iteration, with an average delivery time of 18.4 seconds and a standard deviation of 1.9 seconds. The placement accuracy was consistently high, with an average deviation of 0.1mm from the target position.\\

\noindent {\textbf{Overall System Performance}}
The integrated system's performance across all 10 iterations was evaluated based on the cumulative time taken to complete all five phases, the accuracy of each task, and the overall reliability. The average total cycle time per iteration was recorded as 8.3 minutes, which includes the detection, picking, placing, cutting, unfolding, and delivery processes. The system demonstrated a high level of reliability, with no critical failures observed throughout the experiments.

The results from Table IV emphasize the interdependence of the robot’s actions: a failure in detection directly impacts the picking and placing activities. For instance, in test 7, despite achieving complete success in picking and placing, full cycle success could not be attained due to detection failures. This underlines the importance of reliable detection to ensure overall process success. 
A supplementary video with the whole experimental procedure can be found at this link: \url{https://youtu.be/MXxTeyBVJWg}. 

The successful testing and validation of the proposed solution in the laboratory environment with the FRANKA robot arm showcases its potential for widespread industrial applications. 
The system effectively automated the unpacking and cutting of transparent plastic bags for an 8-stack bulk-loader, meeting the specific requirements and demonstrating robustness under rigorous testing conditions. These results highlight the system's capability to enhance efficiency and safety in industrial processes, aligning with the Industry 4.0 paradigm.


\begin{table*}[h!]
    \centering
    \caption{Pick-n-Place testing for one feeding cycle (8 bags)}
    \setlength{\arrayrulewidth}{0.5mm}
    \setlength{\tabcolsep}{11pt}
    \begin{tabular}{p{0.5cm} p{1.2cm} p{1.2cm} p{1.2cm} p{1.0cm} p{1.2cm} p{1.2cm} p{1.2cm}}
        \toprule
        \textbf{Test} &  \textbf{No. Stacks Successfully Detected} &  \textbf{No. Stacks Successfully Picked} &  \textbf{No. Stacks Successfully Placed} & \textbf{Total Time (min)} & \textbf{Detection Success Rate} (\%) &  \textbf{Picking Success Rate (\%)} & \textbf{Placing Success Rate (\%)} \\
        \midrule
        1 &  8 &  7 &  5 &  8 &  100.00 & 87.50 & 62.50 \\
        2 &  8 &  8 &  8 &  9 &  100.00 & 100.00 & 100.00 \\
        3 &  7 &  5 &  5 &  8 &  87.50 & 62.50 & 62.50 \\
        4 &  8 &  6 &  6 &  8 &  100.00 & 75.00 & 75.00 \\
        5 &  8 &  8 &  8 &  8 &  100.00 & 100.00 & 100.00 \\
        6 &  8 &  7 &  6 &  9 &  100.00 & 87.50 & 75.00 \\
        7 &  6 &  6 &  6 &  8 &  75.00 & 75.00 & 75.00 \\
        8 &  8 &  8 &  8 &  8 &  100.00 & 100.00 & 100.00 \\
        9 &  8 &  6 &  6 &  8 &  100.00 & 75.00 & 75.00 \\
        10 &  8 &  8 &  8 &  9 &  100.00 & 100.00 & 100.00 \\
        \bottomrule
        &  8 & 7 & 7 & 8.3 & 96.25 & 86.25 & 82.5\\
        \hline        
    \end{tabular}
    \label{tab:Pick_and_place}
\end{table*}

\section{DISCUSSION \& CONCLUSIONS}

In this paper, we have presented a comprehensive and innovative system for the autonomous cutting and unpacking of transparent plastic bags in industrial setups, aligned with the principles of Industry 4.0. Leveraging advanced Machine Learning algorithms, particularly CNNs, our system successfully identifies and manipulates transparent plastic bags using a collaborative robot arm equipped with a custom suction cup gripper and depth sensing technologies. The cutting process is facilitated by a combination of vacuum-driven suction cup grippers and a precise linear belt-driven scalpel blade. The delivery system, employing linear actuators and custom door mechanisms, ensures the smooth transition of unpackaged stacks into the bulk loader.

Despite the achievements of our system, there are avenues for further exploration and improvement. Future work could involve enhancing the system's robustness in handling variations in lighting and background conditions, refining the accuracy of detection algorithms, and extending the capabilities of the cutting mechanism to accommodate different types of packaging materials. Integration with more sophisticated artificial intelligence techniques and adaptive control strategies may contribute to further autonomy and efficiency in the unpacking and cutting process. Additionally, exploring the scalability of the system for diverse industrial applications and evaluating its performance in real-world scenarios will be crucial for its widespread adoption. Continuous refinement and adaptation to evolving technologies will be essential to maximize the system's potential in the dynamic landscape of industrial automation.













\bibliographystyle{IEEEtran}
\bibliography{ieeetf}

\end{document}